\documentclass[11pt, oneside]{article}

\usepackage{geometry}
\usepackage{graphicx}
\usepackage{color}
\usepackage{xcolor}
\usepackage{amssymb}
\usepackage{hyperref}
\usepackage{amsmath}
\hypersetup{
    colorlinks=true,
    linkcolor=blue,
    filecolor=magenta,      
    urlcolor=blue,
}
\usepackage{todonotes}
\usepackage{bm}
\usepackage{graphicx}
\usepackage{booktabs}
\usepackage{color, colortbl}
\usepackage{lipsum}
\usepackage{multirow}

\usepackage{subcaption}

\title{VisDA-2021 Competition: \\ Universal Domain Adaptation to Improve Performance on Out-of-Distribution Data}

\author{Dina Bashkirova$^{1}$\thanks{Equal Contribution} \and   Dan Hendrycks$^{3}$\footnotemark[1] \and Donghyun Kim$^{1}$\footnotemark[1]  \and  Samarth Mishra$^{1}$\footnotemark[1]  \and Kate Saenko$^{1,2}$\footnotemark[1]   \and Kuniaki Saito$^{1}$\footnotemark[1]  \and  Piotr Teterwak$^{1}$\footnotemark[1]  \and Ben Usman$^{1}$\footnotemark[1] 
\\ \and
$^{1}$Boston University ~~~~ $^{2}$MIT-IBM Watson AI ~~~~  $^{3}$UC Berkeley \\ \\
 {\tt visda-organizers@googlegroups.com}
}

\date{}

\begin{document}
\maketitle

\begin{abstract}
Progress in machine learning is typically measured by training and testing a model on the same distribution of data, i.e., the same domain. This over-estimates future accuracy on out-of-distribution data. The Visual Domain Adaptation (VisDA) 2021 competition tests models' ability to adapt to novel test distributions and handle distributional shift. We set up unsupervised domain adaptation challenges for image classifiers and will evaluate adaptation to novel viewpoints, backgrounds, modalities and degradation in quality. Our challenge draws on large-scale publically available datasets but constructs the evaluation across domains, rather that the traditional in-domain benchmarking. Furthermore, we focus on the difficult ``universal" setting where, in addition to input distribution drift, methods may encounter missing and/or novel classes in the target dataset. Performance will be measured using a rigorous protocol, comparing to state-of-the-art domain adaptation methods with the help of established metrics. We believe that the competition will encourage further improvement in machine learning methods' ability to handle realistic data in many deployment scenarios. See \url{ http://ai.bu.edu/visda-2021/} 

\end{abstract}

\section{Introduction}

\begin{figure}
    \centering
    \includegraphics[width=0.7\textwidth]{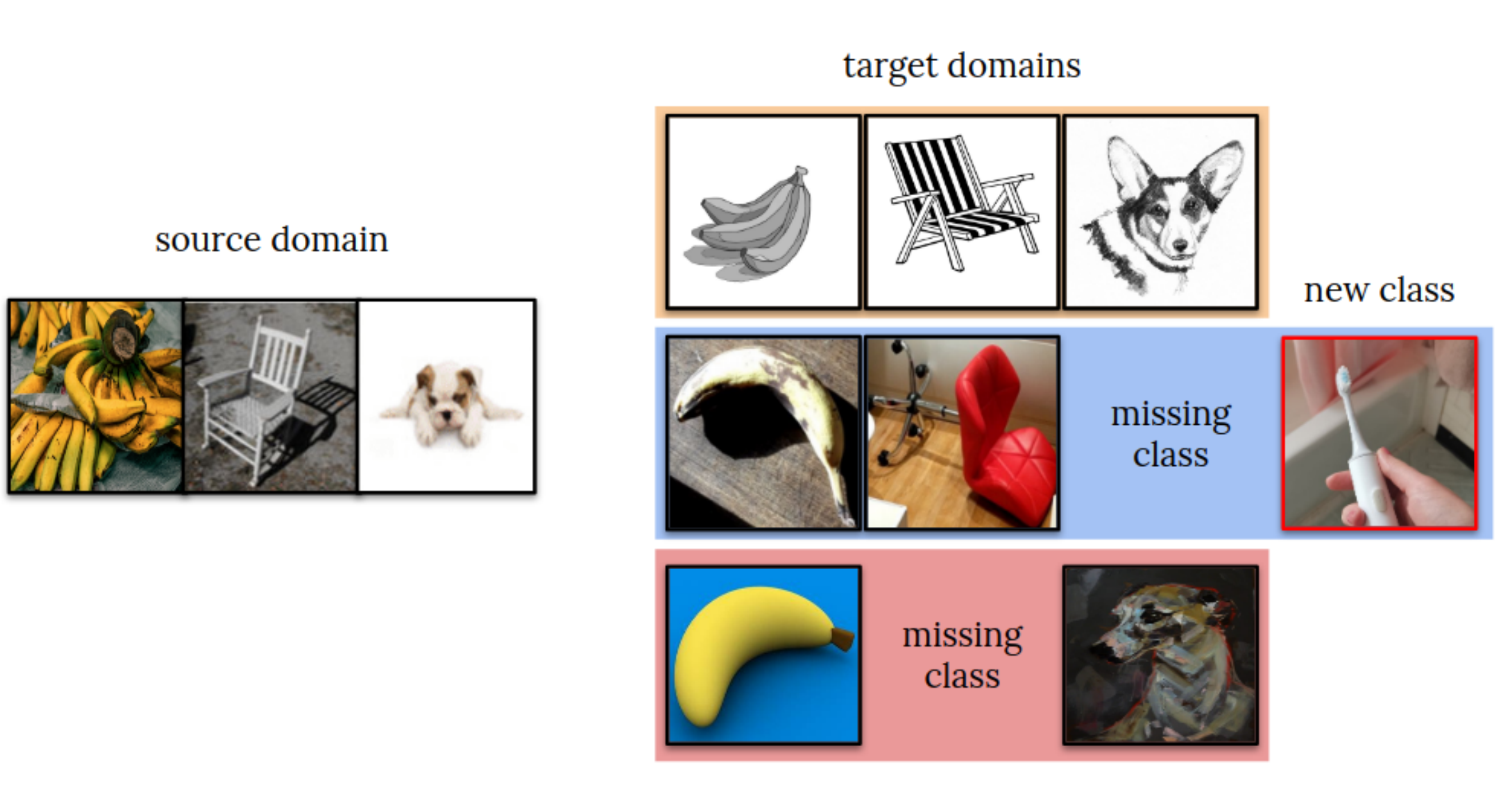}
    \caption{ An illustration of the \textbf{Universal Domain Adaptation} challenge posed by our competition. Given a labeled dataset (source) and an unlabeled dataset (target), the task is to achieve the best possible performance on the target dataset. ``Universal'' refers to fact that, in addition to the input distribution shift, there will be a category shift between the source and target domains. Specifically, the exact category overlap between source and target will be unknown a priori and may include: all classes being shared, missing classes in source, missing classes in target, or both~\cite{saito2020universal}.}
    \label{fig:unida}
\end{figure}

In machine learning, “dataset bias” happens when the training data is not representative of future test data. Finite datasets cannot include all variations possible in the real world, so every machine learning dataset is biased in some way. Yet, machine learning progress is traditionally measured by testing on in-distribution data: almost every new approach is trained on an i.i.d. subset and evaluated on another i.i.d. subset from the same original dataset.

This traditional evaluation obscures the real danger that models will fail on new data distributions. For example, a pedestrian detector trained on pictures of people in the sidewalk could fail on jaywalkers, if the original data collection omitted them. A medical classifier could fail on data from a new hospital or a slightly different patient population. While deep neural networks have significantly improved performance on recognition tasks~\cite{imagenet, simonyan2014very, krizhevsky2012imagenet, faster, maskrcnn}, they still suffer from poor generalization to out-of-domain data~\cite{tzeng2014deep}. 

\textbf{Domain adaptation} techniques aim to evaluate and improve performance on out-of-domain data. The \textbf{Unsupervised Domain Adaptation (UDA)} setting transfers models from a label-rich source domain to an unlabeled target domain without additional supervision. 
Recent UDA methods  achieve this through unsupervised learning on the target domain, e.g., by minimizing the feature distribution shift between source and target domains~\cite{ganin2014unsupervised, long2015learning, sun2015return}, classifier confusion~\cite{jin2020minimum}, clustering~\cite{saito2020universal}, and pseudo-label based methods~\cite{zou2018unsupervised}.
Promising UDA results have been demonstrated on  image classification~\cite{long2017conditional,zou2019confidence, chen2019transferability,xu2019larger,tang2020unsupervised},
 semantic segmentation~\cite{hoffman2016fcns} and object detection~\cite{dafaster} tasks.
 
Our  competition asks teams to solve the \textit{\bf Universal Domain Adaptation (UniDA)} task on visual data. The task is as follows: given a labeled dataset (source) and an unlabeled dataset (target), achieve the best possible performance on the target dataset. ``Universal'' refers to fact that, in addition to the input distribution shift, there is an unspecified category overlap between the source and target domains. Specifically, the exact category overlap between source and target will be unknown a priori and may include: full overlap, missing classes in source, missing classes in target, or both~\cite{saito2020universal}, as illustrated in Fig.~\ref{fig:unida}. \textbf{Robustness} or \textbf{Domain Generalization} is the task of performing well on a target domain without training on it. Common methods include clever augmentations of the training data \cite{hendrycks2020many, hendrycks*2020augmix} and image stylization \cite{geirhos2018imagenet}.  

\subsection{Motivation} 
The domain adaptation problem is highly relevant because generalization to out-of-distribution data is a crucial problem at the heart of machine learning. Humans can generalize across a wide variety of domains, but ML algorithms tend to fail on anything that does not look like their training data. Yet the vast majority of algorithms are evaluated in-distribution, hiding this Achilles' heel of supervised learning. While the domain adaptation sub-field does evaluate the performance of algorithms on out-of-sample data, and has a long history as a research community, the task is still somewhat niche. 

Traditional unsupervised domain adaptation (UDA) methods  assume that all source categories are present in the target domain, however, in practice, little might be known about the category overlap between the two domains.  Our competition is the first to address this more universally applicable UniDA setting. 


\subsection{Applications} 
Distribution shift occurs in many, if not all, real-life application scenarios. When models are trained on offline datasets and deployed in the real world (e.g., a self-driving car, a hospital, etc.) the kinds of data they see inevitably shifts and changes relative to the static training distribution. Collecting more training data in each new domain is not always feasible and interferes with the operation of the system (e.g., imagine having to constantly take it offline to gather more data.) Furthermore, even very large labeled datasets like ImageNet have been found to have a distribution shift from similar datasets collected in the same way~\cite{recht2019}, so dataset bias is always going to be a problem with finite datasets. 

Considering the ubiquity of the problem, there is a strong need for machine learning algorithms that generalize beyond their training data, or at least have the ability to adapt to novel distributions without requiring human supervision. Such algorithms would significantly impact many applications of machine learning. Since our competition focuses on visual data (although the problem also occurs in other modalities), we can imagine several real-world applications where our task is useful: autonomous vehicles navigating in a new environment, a robot encountering objects in the real world and dealing with changing pose, lighting and other factors, or a medical imaging application receiving data from a novel facility or sensor.








\section{Related Competitions}

Members of our team have organized a similar Visual Domain Adaptation (VisDA) challenge at several computer vision conferences in recent years:
\begin{itemize}
    \item The \href{ http://ai.bu.edu/visda-2017/} {1st VISDA (2017)} challenge aimed to test domain adaptation methods’ ability to transfer source knowledge and adapt it to novel target domains, focusing on Sim2Real transfer from a synthetic source domain to a real target domain. It featured an object classification and a semantic image segmentation track.
    \item The \href{http://ai.bu.edu/visda-2018/} {2nd VISDA (2018)} also tackled the Sim2Real problem, but featured object detection and open-set classification tracks.
    \item The \href{http://ai.bu.edu/visda-2019/} {3rd VISDA (2019)} promoted the multi-source and the semi-supervised domain adaptation settings on a newly collected 6-domain DomainNet dataset~\cite{peng2018moment} (real, clipart, painting, drawing, infograph and sketch domains).
    \item The \href{http://ai.bu.edu/visda-2020/} {4th VISDA (2020)} focused on domain adaptive instance retrieval, where the source and target domains have completely different classes (instance IDs), for example, pedestrian IDs. 
\end{itemize}
These challenges were very well received by the computer vision community, with many teams participating and advancing the state-of-the-art in the process. However, the proposed UniDA competition would be the first domain adaptation competition at NeurIPS (or any other machine learning conference) as far as we know. Also, this time we will use different datasets that put emphasis on different types of realistic distribution shifts. The task will also be different: universal domain adaptation (UniDA) where we have unknown overlap between the source and target classes. This more challenging and realistic setting will advance the state-of-the-art in domain transfer even further towards being able to run off the shelf.

There have been several related competitions at NeurIPS. \href{https://www.4paradigm.com/competition/nips2018}{AutoML for Lifelong Machine Learning (NeurIPS'18 competition)} addressed concept drift in lifelong learning, which is different from unsupervised domain adaptation. \href{https://sites.google.com/view/inclusiveimages/}{Inclusive Images} evaluated classification on images drawn from geographic regions underrepresented in the training data. However they provided a labeled validation set from the target distribution, while our competition is focused on truly 'novel' distributions for which no labels were provided. Also, their distributional shift was more narrowly defined as a change in the geographic location where the image was collected by a user (via a crowdsourcing app).

\href{https://sites.google.com/view/pgdl2020}{Predicting Generalization in Deep Learning (NeurIPS'20 competition)} invited competitors to design metrics that accurately predict the generalization performance of deep neural networks without using a test set. The competitors were asked to implement a function that takes a trained model and its training data, and returns a single scalar that correlates with the generalization ability of the model. Our competition is somewhat related in its desire to measure generalization, but addresses a very different task. We do not predict a given models' performance, but evaluate it's actual performance. We also consider generalization to novel domains rather than the same distribution of data. 


\section{VisDA-21 Competition}
\subsection{Datasets}

In the competition, we use ImageNet~\cite{russakovsky2015imagenet}  as our source data. This is a large-scale annotated dataset. contains 1.4M images from 1,000 categories collected from Web. ImageNet  While deep learning models works well on the test set in ImageNet, these can learn biased representation to wrong textures ques~\cite{geirhos2018imagenet} and often do not perform well on data which contains domain-shift including the changes in artistic visual styles, viewpoints, illumination, corruptions, and so on.

The target domains in the competition use images from the following publicly available datasets.
\begin{itemize}
    \item ObjectNet~\cite{barbu2019objectnet} contains 50,000 images containing 313 object classes. Only 113 classes out of the 313 classes are overlapped with ImageNet. The dataset is both easier than ImageNet – objects are largely centered and unoccluded – and harder, due to controlled variations in pose, background and viewpoints. 
    
    \item ImageNet-R~\cite{hendrycks2020many} contains 30,000 images of the 200 classes in ImageNet (partial DA). The images contains different visual styles and textures. 
    \item ImageNet-C~\cite{hendrycks2019benchmarking} contains the same validation images with 1,000 classes in ImageNet (closed set DA) but consists of 15 diverse corruption types, such as blur or noise, with different level of severity. 
    \item ImageNet-O~\cite{hendrycks2021natural} is a dataset built for out-of-distribution detection using imagenet models. It contains a set of 2000 images from Imagenet-22K which do not appear in Imagenet-1K, and are typically classified with high confidence by Imagenet-1K classifiers.
    
\end{itemize}

Under the setting of unsupervised domain adaptation, we allow the access to each unlabeled target domain during training. As we tackle the problem of universal domain adaptation, the goal is to provide a single unified method which performs well on all cross dataset scenarios including full class overlap (closed set), some missing classes in the target (partial set), some novel classes in the target (open set), and a mix of these (open-partial domain adaptation).

We expect these domain shifts to be difficult and will adjust the number of classes accordingly to make the tasks feasible. For example, when tested on ObjectNet, detectors trained on ImageNet lose 40-45\% in performance~\cite{barbu2019objectnet}.

In addition to the above datasets, we  provide development datasets so that teams can tune model architectures and hyperparameters without touching the test datasets. The development datasets are similar to the test datasets in that they are mix of publicly available datasets; however the mixing ratios are different to prevent overfitting as a viable strategy. 




\subsection{Task}
The objective of the competition is to leverage labeled data in the source domain and unlabeled data from the target domain to classify these target domain examples. Keeping with the universal domain adaptation problem, both the source and target domains will have classes that are not found in the other domain. For the images from classes only in the target domain, the task is to detect them as a ``novel'' class not found in the source domain (see Sec. \ref{sec:metrics} for evaluation metrics).

For development/validation, teams are provided a mix of images with labels from ObjectNet, ImageNet-R, ImageNet-C and ImageNet-O. A disjoint set of images from these datasets is used as the test set. In addition, the test set also includes images corrupted via methods different from the ones used in ImageNet-C. A description of these corruptions would be disclosed after the competition concludes.

Teams are asked to submit their predictions on the validation set to find their position on a competition leaderboard based on the metrics described in Sec. \ref{sec:metrics}. Finally towards the end of the competition, teams will be asked to submit their predictions on the test set, for which labels would not be available to the teams. The final leaderboard will be decided by the metrics on this set.

\subsection{Metrics} \label{sec:metrics}
We will use two metrics to evaluate performance during the challenge.

\subsubsection{AUC} 
One desirable characteristic of the model is to be able identify which samples in the evaluation data belong to \textit{novel} classes, in other words those which were not present in the source data. To do this, we ask the the competitiors to give a scalar \textit{novelty score} and compute Area Under the ROC Curve (AUC). The AUC measures unknown (outlier) category detection and is the area under the ROC curve, which is computed by thresholding a score that represents how likely the input is to be an unknown class.

\subsubsection{Instance-wise Accuracy}
We not only want to identify novel classes as OOD data, we also want to have high accuracy on classes present in the source data. Thus, we also employ the instance-wise accuracy over all samples to better understand the behavior of models.

\subsection{Development Kit}
The competition provides a development kit (see \url{ http://ai.bu.edu/visda-2021/}.) The kit contains links to the training, development and test data (test data will be released in the test phase). It also provides baseline code and instructions on how to submit results to the evaluation server, as well as the implementation of metrics.

\subsection{Protocol}

The competition will have two phases. During phase I, the training data (ImageNet as a source domain) and validation domains (a mixture of ObjectNet, ImageNet-R, ImageNet-C and ImageNet-O as target domains) will be available to participants to aid in model development and hyperparameter validation. While we allow participants to use labels for validation domains, high performance on validation domains does not guarantee high performance on test data. During phase II, the test data will be made available for a brief period (two weeks) so that participants can train their models on test data (no labels available for test data) and submit label predictions to the evaluation server. 

\textbf{Tuning on Test Data.} An important part of the protocol is that tuning hyper-parameters or stopping criteria on the target test labels will NOT be allowed. This is because the target data is unsupervised, so supervised tuning is not possible. Therefore successful approaches will have come up with ways to tune without target-domain labels.

\section{Competition Results}

This section will be updated with the winning methods and results after the competition's completion.

\bibliographystyle{ieee}
\bibliography{paper}

\begin{thebibliography}{10}\itemsep=-1pt

\bibitem{barbu2019objectnet}
A.~Barbu, D.~Mayo, J.~Alverio, W.~Luo, C.~Wang, D.~Gutfreund, J.~Tenenbaum, and
  B.~Katz.
\newblock Objectnet: A large-scale bias-controlled dataset for pushing the
  limits of object recognition models.
\newblock {\em Advances in neural information processing systems},
  32:9453--9463, 2019.

\bibitem{chen2019transferability}
X.~Chen, S.~Wang, M.~Long, and J.~Wang.
\newblock Transferability vs. discriminability: Batch spectral penalization for
  adversarial domain adaptation.
\newblock 2019.

\bibitem{dafaster}
Y.~Chen, W.~Li, C.~Sakaridis, D.~Dai, and L.~Van~Gool.
\newblock Domain adaptive faster r-cnn for object detection in the wild.
\newblock 2018.

\bibitem{imagenet}
J.~Deng, W.~Dong, R.~Socher, L.-J. Li, K.~Li, and L.~Fei-Fei.
\newblock Imagenet: A large-scale hierarchical image database.
\newblock 2009.

\bibitem{ganin2014unsupervised}
Y.~Ganin and V.~Lempitsky.
\newblock Unsupervised domain adaptation by backpropagation.
\newblock 2014.

\bibitem{geirhos2018imagenet}
R.~Geirhos, P.~Rubisch, C.~Michaelis, M.~Bethge, F.~A. Wichmann, and
  W.~Brendel.
\newblock Imagenet-trained cnns are biased towards texture; increasing shape
  bias improves accuracy and robustness.
\newblock {\em arXiv preprint arXiv:1811.12231}, 2018.

\bibitem{maskrcnn}
K.~He, G.~Gkioxari, P.~Doll{\'a}r, and R.~Girshick.
\newblock Mask r-cnn.
\newblock 2017.

\bibitem{hendrycks2020many}
D.~Hendrycks, S.~Basart, N.~Mu, S.~Kadavath, F.~Wang, E.~Dorundo, R.~Desai,
  T.~Zhu, S.~Parajuli, M.~Guo, et~al.
\newblock The many faces of robustness: A critical analysis of
  out-of-distribution generalization.
\newblock {\em arXiv preprint arXiv:2006.16241}, 2020.

\bibitem{hendrycks2019benchmarking}
D.~Hendrycks and T.~Dietterich.
\newblock Benchmarking neural network robustness to common corruptions and
  perturbations.
\newblock {\em arXiv preprint arXiv:1903.12261}, 2019.

\bibitem{hendrycks*2020augmix}
D.~Hendrycks*, N.~Mu*, E.~D. Cubuk, B.~Zoph, J.~Gilmer, and
  B.~Lakshminarayanan.
\newblock Augmix: A simple method to improve robustness and uncertainty under
  data shift.
\newblock In {\em International Conference on Learning Representations}, 2020.

\bibitem{hendrycks2021natural}
D.~Hendrycks, K.~Zhao, S.~Basart, J.~Steinhardt, and D.~Song.
\newblock Natural adversarial examples.
\newblock In {\em Proceedings of the IEEE/CVF Conference on Computer Vision and
  Pattern Recognition}, pages 15262--15271, 2021.

\bibitem{hoffman2016fcns}
J.~Hoffman, D.~Wang, F.~Yu, and T.~Darrell.
\newblock Fcns in the wild: Pixel-level adversarial and constraint-based
  adaptation.
\newblock {\em arXiv preprint arXiv:1612.02649}, 2016.

\bibitem{jin2020minimum}
Y.~Jin, X.~Wang, M.~Long, and J.~Wang.
\newblock Minimum class confusion for versatile domain adaptation.
\newblock 2020.

\bibitem{krizhevsky2012imagenet}
A.~Krizhevsky, I.~Sutskever, and G.~E. Hinton.
\newblock Imagenet classification with deep convolutional neural networks.
\newblock 2012.

\bibitem{long2015learning}
M.~Long, Y.~Cao, J.~Wang, and M.~I. Jordan.
\newblock Learning transferable features with deep adaptation networks.
\newblock 2015.

\bibitem{long2017conditional}
M.~Long, Z.~Cao, J.~Wang, and M.~I. Jordan.
\newblock Conditional adversarial domain adaptation.
\newblock 2018.

\bibitem{peng2018moment}
X.~Peng, Q.~Bai, X.~Xia, Z.~Huang, K.~Saenko, and B.~Wang.
\newblock Moment matching for multi-source domain adaptation.
\newblock 2019.

\bibitem{recht2019}
B.~Recht, R.~Roelofs, L.~Schmidt, and V.~Shankar.
\newblock Do imagenet classifiers generalize to imagenet?
\newblock {\em CoRR}, abs/1902.10811, 2019.

\bibitem{faster}
S.~Ren, K.~He, R.~Girshick, and J.~Sun.
\newblock Faster r-cnn: Towards real-time object detection with region proposal
  networks.
\newblock 2015.

\bibitem{russakovsky2015imagenet}
O.~Russakovsky, J.~Deng, H.~Su, J.~Krause, S.~Satheesh, S.~Ma, Z.~Huang,
  A.~Karpathy, A.~Khosla, M.~Bernstein, et~al.
\newblock Imagenet large scale visual recognition challenge.
\newblock {\em International journal of computer vision}, 115(3):211--252,
  2015.

\bibitem{saito2020universal}
K.~Saito, D.~Kim, S.~Sclaroff, and K.~Saenko.
\newblock Universal domain adaptation through self supervision.
\newblock {\em arXiv preprint arXiv:2002.07953}, 2020.

\bibitem{simonyan2014very}
K.~Simonyan and A.~Zisserman.
\newblock Very deep convolutional networks for large-scale image recognition.
\newblock {\em arXiv}, 2014.

\bibitem{sun2015return}
B.~Sun, J.~Feng, and K.~Saenko.
\newblock Return of frustratingly easy domain adaptation.
\newblock In {\em AAAI}, 2016.

\bibitem{tang2020unsupervised}
H.~Tang, K.~Chen, and K.~Jia.
\newblock Unsupervised domain adaptation via structurally regularized deep
  clustering.
\newblock 2020.

\bibitem{tzeng2014deep}
E.~Tzeng, J.~Hoffman, N.~Zhang, K.~Saenko, and T.~Darrell.
\newblock Deep domain confusion: Maximizing for domain invariance.
\newblock {\em arXiv}, 2014.

\bibitem{xu2019larger}
R.~Xu, G.~Li, J.~Yang, and L.~Lin.
\newblock Larger norm more transferable: An adaptive feature norm approach for
  unsupervised domain adaptation.
\newblock 2019.

\bibitem{zou2019confidence}
Y.~Zou, Z.~Yu, X.~Liu, B.~Kumar, and J.~Wang.
\newblock Confidence regularized self-training.
\newblock 2019.

\bibitem{zou2018unsupervised}
Y.~Zou, Z.~Yu, B.~Vijaya~Kumar, and J.~Wang.
\newblock Unsupervised domain adaptation for semantic segmentation via
  class-balanced self-training.
\newblock 2018.

\end{thebibliography}
\section*{Appendix: Past Related Workshops and Tutorials}

\begin{enumerate}
    \item NIPS2011, Workshop on \href{https://sites.google.com/site/nips2011domainadap}{\textit{``Domain Adaptation: Theory and Application''}}. 
    \item NIPS2011, Workshop on \href{https://sites.google.com/site/nips2011workshop/}{\textit{Challenges in Learning Hierarchical Models: Transfer Learning and Optimization}}
    \item CVPR2012, half day Tutorial on \href{http://vc.sce.ntu.edu.sg/transferlearning.html}{\textit{Domain Transfer Learning for Vision Applications}}
    \item IJCV2013, Special Issue on \href{http://lists.diku.dk/pipermail/imageworld/2013-February/005247.html}{\textit{Domain Adaptation for Vision Applications}}
    \item ICCV2013, Workshop on \href{http://visda2013.seas.harvard.edu}{\textit{Visual Domain Adaptation and Dataset Bias (VisDA)''}}
    \item NIPS2013, Workshop on \href{https://sites.google.com/site/learningacross}{\textit{New Directions in Transfer and Multi-Task: Learning Across Domains and Tasks}}
    \item ECCV2014, Workshop on \href{http://www.cvc.uab.es/adas/task-cv2014/}{\textit{Transferring and Adapting Source Knowledge (TASK) in Computer Vision (CV)}}
    \item ECCV2014, Tutorial on\href{http://tommasit.wix.com/datl14tutorial}{\textit{Domain Adaptation and Transfer Learning}}
    \item ECML2014, Workshop on \href{http://users.dsic.upv.es/~flip/LMCE2014/}{\textit{Learning over Multiple Contexts.}}
    \item NIPS2014, Workshop on \href{https://sites.google.com/site/multitaskwsnips2014/}{\textit{Transfer and Multi-Task Learning: Theory Meets Practice.}}
    
    \item JMLR2015, Special Issue on \href{https://sites.google.com/site/jmlrmtldatl/}{\textit{Multi-Task Learning, Domain Adaptation and Transfer Learning}}
    
    \item ICDM2015, Workshop on \href{https://sites.google.com/site/icdmwptl2015/home}{Practical Transfer Learning}
    
    \item ICCV2015, Workshop on \href{http://adas.cvc.uab.es/task-cv2015/}{\textit{Transferringand Adapting Source Knowledge (TASK) in Computer Vision (CV)}}
    
    \item NIPS2015, Workshop on \href{https://sites.google.com/site/tlworkshop2015/}{\textit{Transfer and Multi-Task Learning: Trends and New Perspectives}}
    
    \item ECCV2016, Workshop on \href{http://adas.cvc.uab.es/task-cv2016/}{\textit{Transferring and Adapting Source Knowledge (TASK) in Computer Vision (CV)}}
    
    \item ICCV2017, Workshop on \href{http://adas.cvc.uab.es/task-cv2017/}{\textit{Transferring and Adapting Source Knowledge (TASK) in Computer Vision (CV)}}
    
    \item BigData2017, Workshop on \href{https://easychair.org/cfp/BDTL_Workshop_2017}{\textit{Big Data Transfer Learning (BDTL)}}
    
    \item AAMAS2017, Workshop on \href{http://www.tirl.info/}{\textit{Transfer in Reinforcement Learning (TiRL)}}
 
    \item ACCV 2018, Workshop on \href{http://users.cecs.anu.edu.au/~koniusz/openmic-accv18/}{\textit{Domain Adaptation and Few-Shot Learning (Open MIC)}}
    
    \item ECCV2018, Workshop on \href{https://sites.google.com/view/task-cv2018/home}{\textit{Transferring and Adapting Source Knowledge (TASK) in Computer Vision (CV)}}
    
    \item ICIAP2019, Tutorial on \href{https://event.unitn.it/iciap2019/#tutorials}{\textit{Transferring Knowledge Across Domains: an Introduction to Deep Domain Adaptation}};
    
    \item ICCV2019, tutorial on \href{ https://sites.google.com/view/learning-with-limited-data/}{\textit{Visual Learning with Limited Labeled Data}}.
    
    \item ICCV2019, Workshop on \href{https://sites.google.com/view/task-cv2019/home}{\textit{Transferring and Adapting Source Knowledge (TASK) in Computer Vision (CV)}}

    \item ECCV2020, Workshop on \href{https://sites.google.com/view/task-cv2020/home}{\textit{Transferring and Adapting Source Knowledge (TASK) in Computer Vision (CV)}}
    
\end{enumerate}

\end{document}